\documentclass{article}
\usepackage{spconf,amsmath,epsfig,multirow,adjustbox,amssymb,makecell,tabularx, hyperref} 
\usepackage{bm}
\usepackage[T1]{fontenc}
\usepackage[utf8]{inputenc}

\title{Decoupling Localization and Classification in Single Shot Temporal Action Detection}
\name{Yupan Huang$^1$\sthanks{Work done during an internship at Microsoft Research Asia.}, Qi Dai$^2$, and Yutong Lu$^1$\sthanks{Corresponding Author.}}
\address{$^1$ Sun Yat-sen University, Guangzhou, China \\
	$^2$ Microsoft Research Asia, Beijing, China \\
	\small\texttt{huangyp28@mail2.sysu.edu.cn}, \texttt{qid@microsoft.com}, \texttt{yutong.lu@nscc-gz.cn}}
\begin{document}\sloppy
\maketitle
\begin{abstract}
Video temporal action detection aims to temporally localize and recognize the action in untrimmed videos. 
Existing one-stage approaches mostly focus on unifying two subtasks, i.e., localization of action proposals and classification of each proposal through a fully shared backbone. However, such design of encapsulating all components of two subtasks in one single network might restrict the training by ignoring the specialized characteristic of each subtask.
In this paper, we propose a novel Decoupled Single Shot temporal Action Detection (Decouple-SSAD) method to mitigate such problem by decoupling the localization and classification in a one-stage scheme.
Particularly, two separate branches are designed in parallel to enable each component to own representations privately for accurate localization or classification.
Each branch produces a set of action anchor layers by applying deconvolution to the feature maps of the main stream.
High-level semantic information from deeper layers is thus incorporated to enhance the feature representations.
We conduct extensive experiments on THUMOS14 dataset and demonstrate superior performance over state-of-the-art methods.
Our code is available online\footnote{https://github.com/hypjudy/Decouple-SSAD}.
\end{abstract}
\section{Introduction}
\label{sec:intro}
With the tremendous increase in online and personal media archives, people are generating, storing, and consuming a large collection of videos.
This trend encourages the development of effective and efficient algorithms to intelligently
parse video data \cite{simonyan2014two,qiu2017deep,li2016action,qiu2017learning,li2018unified,qiu2019learning} and discover semantic information \cite{li2018jointly,qiu2018learning}.
One fundamental challenge underlying the success of these advances is action detection from videos in both temporal \cite{shou2016temporal,long2019gaussian} and spatio-temporal aspects \cite{li2018recurrent}.
In this study, we focus on the temporal action detection task, which aims to find the exact time stamps of an action's start and end time, and recognize the category of the action.

Recent works have been inspired by object detection, which can be categorized as either two-stage approach \cite{wang2017untrimmednets,xu2017r,chao2018rethinking,lin2018bsn} or one-stage approach \cite{lin2017single,yeung2016end,sstad_buch_bmvc17}.
The two-stage approaches follow the framework of first generating action proposals and then classifying them. 
The main drawback of these two-stage approaches is the indirect optimization strategy, which may result in a sub-optimal solution.
Contrarily, the one-stage approaches directly produce and classify the action proposals in only one step, which is an end-to-end framework.
However, such design of encapsulating all components of two subtasks in one single network might restrict the training by ignoring the specialized characteristic of each subtask.
For example, given two different proposals near the same ground truth action instance, the classification optimization aims to pull their feature representations together, while the localization optimization would separate them since they have different offset values.
Both categories of methods would produce unsatisfactory performance.

To solve the above problems, we propose a novel Decoupled Single Shot temporal Action Detection (Decouple-SSAD) method to inherit the advantages from both directions.
Decouple-SSAD improves the one-stage action detection methods by additionally involving two branches, i.e., the refined classification branch and the refined proposal branch. 
The two branches manually separate the proposal generation and classification process, preserving the learning of specialized features useful for optimizing the performance of each component.
Specially, each branch produces a set of action anchor layers by applying deconvolution to the feature maps of the main stream.
High-level semantic information from deeper layers is thus incorporated to enhance the features.
The whole framework is trained in an end-to-end manner.
With such a one-stage framework, Decouple-SSAD could not only optimize the proposal and recognition parts jointly but also provide additional refinement for each part.

The main contribution of this paper is the proposed Decouple-SSAD for addressing the issue of temporal action detection.
The method provides an elegant solution by incorporating the separate optimization for proposal and classification into a one-stage action detection framework. Decouple-SSAD achieves state-of-the-art performance on the standard benchmark THUMOS14, demonstrating its effectiveness.
 
\section{Related Work}
We summarize recent works related to our approach into two categories: object detection and temporal action detection.

\textbf{Object Detection.}
Recent works on temporal action detection \cite{lin2017single,chao2018rethinking} get inspiration from the works on object detection.
These works can be divided into two categories: two-stage approaches (e.g., RCNN \cite{girshick2014rich} and Faster RCNN \cite{ren2015faster}) and one-stage approaches (e.g., YOLO \cite{redmon2016you} and SSD \cite{liu2016ssd}).
Recently, RefineDet \cite{zhang2018single} proposes to refine the object localization in the one-stage framework by transferring information from deep anchor layers to shallow refined layers to predict accurate locations and classes, which is similar to our Decouple-SSAD. However, Decouple-SSAD is more robust as it learns specialized features separately for localization and classification, whereas RefineDet refines the detection with coarse results in each anchor layer.

\textbf{Temporal Action Detection.}
As mentioned previously, existing temporal action detection methods can be divided into two categories: two-stage approaches and one-stage approaches. The two-stage approaches first generate action instances and then classify them. Some works focus on proposal generation \cite{escorcia2016daps,buch2017sst,Gao_2017_ICCV,lin2018bsn,gao2018ctap,AAAI1816109} while others concentrate on classification \cite{xu2017r,dai2017temporal,zhao2017temporal}.
The one-stage approaches integrate proposal and classification into a single step.
SSAD \cite{lin2017single} utilizes 1D convolution to generate multiple temporal anchor action instances for action classification and boundary box regression. The Single Stream Temporal Action Detection (SS-TAD) \cite{sstad_buch_bmvc17} utilizes the RNN-based architecture to learn action proposal and classification jointly.
Yeung et al. \cite{yeung2016end} explored RNN to learn a glimpse policy for predicting the starting and ending points of actions in an end-to-end manner.
Nevertheless, both two-stage and one-stage approaches have drawbacks that the former adopts the indirect optimization and the later usually generates inaccurate proposals.
In this research, we leverage the advantages of both directions and overcome their weaknesses by involving two separate branches into the one-stage framework.

\begin{figure*}[ht]
    \includegraphics[width=\textwidth]{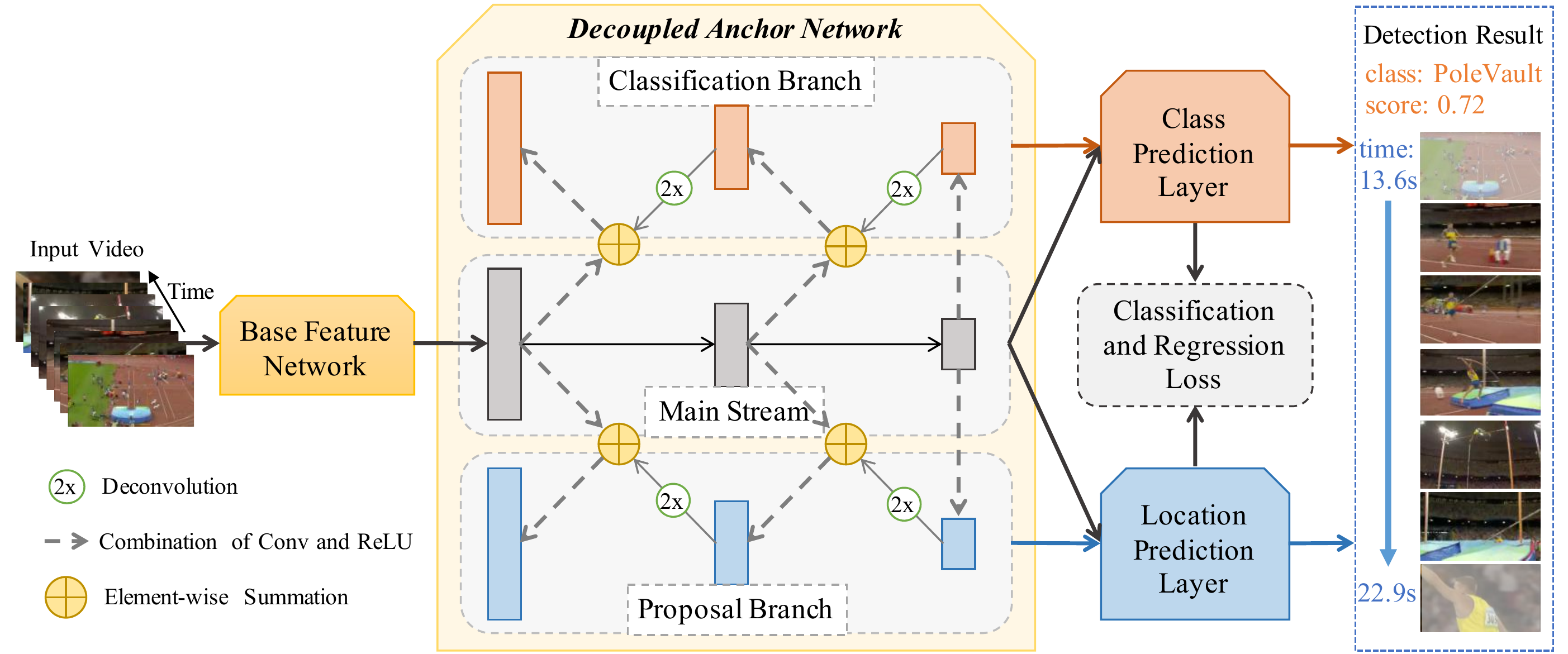}
    \centering
    \caption{\small The framework overview of our Decouple-SSAD. The input frame sequence is first encoded to $1d$ feature sequence through $2d/3d$ ConvNet.
    A multi-branch anchor network is designed to produce multiple sets of feature maps.
    Particularly, it consists of a main stream and two branches derived from it, i.e., the refined classification branch and proposal branch.
    The Main Stream performs classification and regression simultaneously.
    The Classification and Proposal Branches focus on learning specialized features useful for themselves, which incorporate the semantic information of deep layers from main stream through deconvolution.
    The action classification loss and location regression loss are utilized to optimize different branches. The whole network is trained in an end-to-end manner.
    }
    \label{fig:architecture}
\end{figure*}

\section{Approach}
In this section, we present the proposed Decoupled Single Shot temporal Action Detection (Decouple-SSAD) in detail. Figure \ref{fig:architecture} shows an overview of our architecture, which consists of three main components: the base feature network, the anchor network, and the classification
and regression module.
The base feature network extracts representations of each frame or clip for later action detection.
Then a multi-branch anchor network is utilized to localize the proposals and classify them accurately. Specifically, we separate proposal generation and category recognition process by additionally adding two branches to perform proposal and classification refinement respectively. Finally, we exploit the action classification and temporal boundary box regression module for accurate detection. Different types of optimization are employed for each branch.
The whole framework can be trained in an end-to-end manner.
The above design not only optimizes the proposal and recognition parts jointly but also provides additional refinement for each part.

\subsection{Base Feature Network}
Our goal is to detect action segments in $1d$ temporal space. To achieve this, given the video frame sequence, we first extract a 1d feature map for each frame/clip, typically via a $2d$/$3d$ ConvNet. We then pass the sequence of feature maps to the $1d$ ConvNet, converting it into a detection problem in $1d$ domain. In particular, a sequence of frame/clip level CNN features $\{f_i\}^{T-1}_{i=0}$ are extracted, where $T$ is temporal length. We further feed the features into two $1d$ convolutional layers plus one max-pooling layer to increase its receptive field for better detection. We use the output of the base feature network for proposal generation and feature pooling in the next step.

\subsection{Network Architecture}
Given the 1d sequence of feature maps, the common one-stage action detection solution stacks $1d$ anchor convolutional layers of different scales to generate anchor proposals for classification and boundary regression \cite{lin2017single}. However, such one-step scheme encapsulates all components of two subtasks (classification and localization) in a single network, restricting the training by ignoring the specialized characteristic of each subtask. 
To this end, we propose to additionally involve refined proposal and classification branches, which focus on learning specialized features useful for optimizing the performance of each subtask.
Our anchor network thus consists of three parts: the main stream, the classification branch, and the proposal branch.

\textbf{Main Stream.}
The main stream predicts the locations and categories of action instances simultaneously. Based on the feature representations from the base network, it generates a set of anchor layers with different scales by temporal pooling.
Particularly, $N_l$ convolutional layers are cascaded, each with one anchor layer to produce action proposals. Denote the feature map of $j$-th convolutional layer in main stream as $\bm{f}_{m}^j$, $1\leq j\leq N_l$.
In each anchor layer, we produce a set of action proposals with various aspect ratios at each temporal location.
With the feature representation of each anchor action proposal $(a_c, a_w)$ where $a_c$ and $a_w$ are the default center position and width, we utilize $1d$ convolutional layers to output three predictions for action recognition and localization: $i$) action classification scores $\mathbf{p}=[p_0, p_1, ..., p_C]$ indicating the probabilities belong to $C$ action categories plus one ``background'' class, $ii$) the overlap value $p_{ov}$ representing the IoU between the default proposal $(a_c, a_w)$ and its closest ground truth proposal, and $iii$) proposal regression parameters $(\Delta c, \Delta w)$ denoting the center and width offsets for $(a_c, a_w)$.
Finally, we obtain the predicted proposal $(\varphi_c, \varphi_w)$ by
\begin{equation}\label{Eq1}\small
       \begin{split}
              \varphi_c = a_c + \alpha_1 a_w \Delta c ~~~{\rm{and}} ~~~\varphi_w = a_w \exp{(\alpha_2 \Delta w)}~,
       \end{split}
\end{equation}
where $\varphi_c$, $\varphi_w$ are the center position and width, and $\alpha_1$, $\alpha_2$ are the parameters for adjusting the effect of $\Delta c, \Delta w$.

The detection results of the above main stream network are usually unsatisfactory.
On the one hand, it ignores the specialized characteristic of each subtask.
On the other hand, the lower anchor layers lack the rich semantic clues from deep layers for accurate detection.
To address these issues, we introduce two additional branches, that is, the classification branch and the proposal branch.

\textbf{Classification Branch.}
This branch focuses on the classification of proposals. To enhance its performance in lower anchor layers, we propose to incorporate the rich semantic information in deeper layers through deconvolution.
Instead of the direct deconvolution on the main stream to obtain the classification branch, we combine the deconvolutional feature map with one of the shallower layers in the main stream to generate the feature map $\{\bm{f}_c^j\}_{j=1}^{N_l}$ in this branch.
Particularly, the feature map $\bm{f}_c^j$ in $j$-th layer is given by 
\begin{equation}\label{Eq2}\small
\bm{f}_{c}^j =
\begin{cases}
~~~\mathcal{C}_1(\bm{f}_{m}^{j})~~~~~~~~~~~~~~~~~~~~~~~~~~~~~~~,~~~ \text{if~~$j=N_l$} \\
~~~\mathcal{C}_2(\mathcal{S}(\mathcal{C}_3(\bm{f}_{m}^j),~ \mathcal{D}(\bm{f}_{c}^{j+1}))),~~ \text{if~~$1\leq j< N_l$}
\end{cases},
\end{equation}
where $\mathcal{C}_1(\cdot)$, $\mathcal{C}_2(\cdot)$, $\mathcal{C}_3(\cdot)$ are the combinations of convolution and ReLU operations for increasing the discriminative ability, $\mathcal{D(\cdot)}$ indicates the deconvolution, and $\mathcal{S}(\cdot,\cdot)$ denotes the element-wise summation, $\mathcal{S}(\bm{a}, \bm{b})=\rho\cdot\bm{a}+(1-\rho)\cdot\bm{b}$, $\rho$ is the hyper-parameter.
We produce a set of action proposals at each temporal location of anchor layers, as done similarly in the main stream, and predict their action classification scores $\mathbf{p}_c=[\hat{p}_0, \hat{p}_1, ..., \hat{p}_C]$.
In addition, we adopt an average fusion to fuse the scores of the classification branch and the main stream, obtaining new scores $\mathbf{p}_c'$ for classification branch as
\begin{equation}\label{Eq4}\small
\begin{split}
&\mathbf{p}_c' = (\mathbf{p}+\mathbf{p}_c)/2.
\end{split}
\end{equation}
$\mathbf{p}_c'$ is then used for training and testing classification branch.

With the classification branch, we take advantages of both main stream and refined classification branch to recognize actions from coarse to fine.
The main stream first conducts predictions on both location and category scores while the classification branch performs a second refinement on recognition.

\textbf{Proposal Branch.}
This branch concentrates on providing the exact start and end time stamps of action instances in each video.
We propose to involve high-level semantic information into the proposal branch, as done similarly in the classification branch.
In detail, the feature map $\bm{f}_p^{j}$ of $j$-th layer in proposal branch is formulated as follows,
\begin{equation}\label{Eq3}\small
\bm{f}_{p}^j =
\begin{cases}
~~~\mathcal{C}_4(\bm{f}_{m}^{j})~~~~~~~~~~~~~~~~~~~~~~~~~~~~~~~,~~~ \text{if~~$j=N_l$} \\
~~~\mathcal{C}_5(\mathcal{S}(\mathcal{C}_6(\bm{f}_{m}^j),~ \mathcal{D}(\bm{f}_{p}^{j+1}))),~~ \text{if~~$1\leq j< N_l$}
\end{cases},
\end{equation}
where $\mathcal{C}_4(\cdot)$, $\mathcal{C}_5(\cdot)$, $\mathcal{C}_6(\cdot)$ are also the combinations of convolution and ReLU operations.
We then generate a set of anchor layers, in which action proposals are produced by outputting two predictions: the overlap value $p_{ov\_p}$ and proposal offsets $(\Delta c_p,\Delta w_p)$.
The average fusion strategy is also adopted to fuse the two predictions
of the proposal branch and the main stream.
The overlap prediction $p_{ov}'$ and the offset prediction $(\Delta c',\Delta w')$ for proposal branch is thus calculated by
\begin{equation}\label{Eq4}\small
\begin{split}
&p_{ov}' = (p_{ov}+p_{ov\_p})/2, \\
&\Delta c' = (\Delta c+\Delta c_p)/2, \\
&\Delta w' = (\Delta w+\Delta w_p)/2.
\end{split}
\end{equation}
The average fusion strategy incorporates the complementary information from the main stream into the proposal branch and classification branch and could make an improvement of about 3\% mAP@0.5 on THUMOS14.

\subsection{Training and Inference}

\textbf{Loss Function.}
The proposed network is optimized via three types of supervision: classification loss, location regression loss, and overlap loss. Particularly, the classification branch is optimized with classification loss, while the proposal branch is optimized with the other two losses. The main stream is optimized with all three losses.

For classification, we utilize the standard softmax loss, which is formulated as
\begin{equation}\label{Eq5}\small
L_{cls} = -\sum\limits_{n=0}^{C}I_{n=c}\log(p_{n}),
\end{equation}
where $I_{n=c}$ is an indicator function which equals to 1 if $n$ is the ground truth class label $c$, otherwise 0.
For location regression, we employ the Smooth L1 loss ($S_{L1}$) to force the proposal $(\varphi_c, \varphi_w)$ to move towards its closest ground truth proposal $(g_c, g_w)$. The loss is computed by
\begin{equation}\label{Eq6}
       \small
       {L}_{reg} = S_{L1}(\varphi_{c}-g_{c})+S_{L1}(\varphi_{w}-g_{w}).
\end{equation}
The IoU overlap loss is the Mean Square Error (MSE) loss:
\begin{equation}\label{Eq7}
\small
L_{ov} = (p_{ov} - g_{iou})^2,
\end{equation}
where $g_{iou}$ is the ground truth IoU value between the proposal and its closest ground truth.

Finally, we accumulate the losses from three branches. The overall training objective is defined as follows
\begin{equation}\label{Eq8}
       \small
       {L} = \alpha \cdot{L}_{cls} + \beta \cdot{L}_{reg} + \gamma \cdot{L}_{ov},
\end{equation}
where ${L}_{cls}=\omega \cdot L_{cls,m}+(1-\omega) \cdot L_{cls,c}$, $L_{reg}=\omega \cdot L_{reg,m}+(1-\omega) \cdot L_{reg,p}$, $L_{ov}=\omega \cdot L_{ov,m}+(1-\omega) \cdot L_{ov,p}$, and $\alpha,\beta,\gamma,\omega$ are the parameters for balancing different tasks and branches.

\textbf{Inference.}
During testing, the classification scores from classification branch are utilized as the final prediction scores. For localization, as aforementioned, we exploit the fusion of proposal branch and main stream to obtain the final offsets $(\Delta c',\Delta w')$, which are further used to compute the proposals. Finally, we apply the Non-Maximum Suppression (NMS) with Jaccard overlap of 0.2 to produce the detection results.

\section{Experiments}

\begin{figure*}[ht]
    \includegraphics[width=\textwidth]{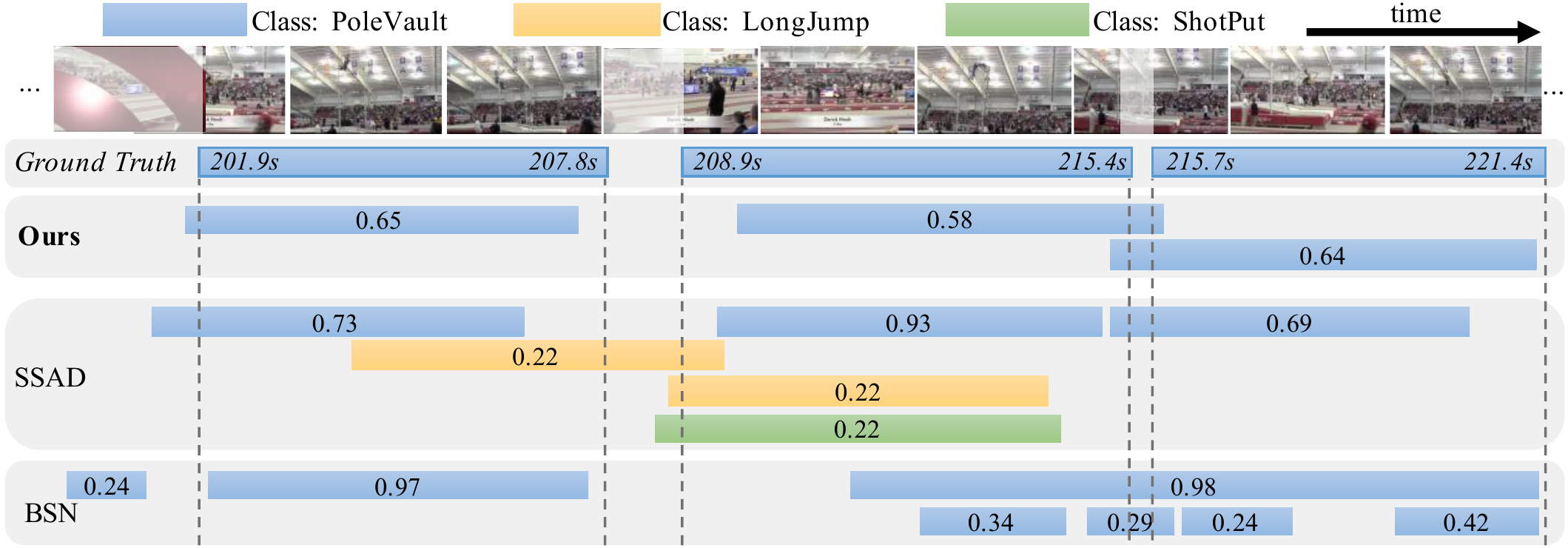}
    \centering
    \caption{Visualization of predicted action instances by our Decouple-SSAD, SSAD, and BSN on one video from THUMOS14. Each bar represents an action instance. The number in the middle of each bar denotes the confidence score of the prediction. Decouple-SSAD provides more accurate category prediction and more precise boundaries than SSAD and BSN. }
    \label{fig:visualization}
\end{figure*}

\subsection{Implementation Details}
\label{implementation}
\textbf{Dataset.}
We conduct experiments on the THUMOS14~\cite{THUMOS14}.
The training set known as UCF101~\cite{soomro2012ucf101} has 13,320 trimmed videos and is often used for pre-training.
It has 1,010 untrimmed validation videos and 1,574 untrimmed test videos of 20 classes. It is a standard practice for researchers to train on the 200 temporal annotated validation set and evaluate on the 213 public test videos.
We report mean Average Precision (mAP) metric with IoU thresholds varied from 0.3 to 0.7.

\textbf{Hard Negative Mining.}
Following \cite{lin2017single}, we adopt the hard negative mining strategy to handle the positive-negative class imbalance problem. 
Negative samples are defined as the instances whose Jaccard overlap values with their closest ground truth instances are smaller than $0.5$, while positive samples are the opposite. Then the hard negative samples are instances with predicted overlaps larger than $0.5$. We select all hard negative samples and let the ratio between negative and positive be 1:1.

\textbf{Network Details.}
We utilize two-stream networks \cite{simonyan2014two} pre-trained on UCF101 \cite{soomro2012ucf101} to extract spatial and temporal feature representations for each clip with length $512$, and feed the feature sequence into our network. We also evaluate the feature extractor pre-trained on Kinetics \cite{kay2017kinetics} when compared with the state-of-the-art.
The number of anchor layers $N_l$ is $3$. The aspect ratios at each temporal location are $\{0.5,0.75,1,1.5,2\}$. We set $\alpha_1$, $\alpha_2$ to $0.1$.
When producing anchor feature maps of the two branches, $\mathcal{C}_1(\cdot)$, $\mathcal{C}_4(\cdot)$ consist of three ``Conv-ReLU'' units, $\mathcal{C}_2(\cdot)$, $\mathcal{C}_5(\cdot)$ denote the ``Conv-ReLU-Conv'' unit, and $\mathcal{C}_3(\cdot)$, $\mathcal{C}_6(\cdot)$ are the ``ReLU-Conv-ReLU'' unit.
We set $\rho=2/3$, $\omega=2/3$, $\alpha=1$, and $\beta=\gamma=10$ through the cross validation.
In all of our experiments, our network is trained by utilizing Adam algorithm. The training process takes 30 epochs with the learning rate of $10^{-4}$. The batch size is 48.

\begin{table}[t!]
\centering
\caption{Effectiveness of different components in Decouple-SSAD. Refinement branch indicates performing the localization and classification refinement in only one branch.} 
\label{tab:two-branch}
\begin{tabular}{ c|c c c c c } 
\Xhline{2\arrayrulewidth}
Component & \multicolumn{5}{c}{Performance}\\
\hline
Main Stream &\checkmark& \checkmark & \checkmark&\checkmark & \checkmark   \\
Classification Branch &&& \checkmark & & \checkmark   \\
Proposal Branch& & \checkmark & & &  \checkmark \\
Refinement Branch && & & \checkmark &  \\
\hline
mAP@0.5 (\%)& 31.2& 32.2 & 33.4& 34.0& \textbf{35.8}    \\
\Xhline{2\arrayrulewidth}
\end{tabular}
\end{table} 

\subsection{Ablation Study}
We conduct an ablation study for analyzing different components of Decouple-SSAD.
The baseline denotes the network with an individual main stream to perform both localization and classification, which is identical to the SSAD \cite{lin2017single} but with different settings.
We make several modifications and improve mAP@0.5 performance of SSAD from 24.6\% to 31.2\%. For more details, please refer to our public code\footnote{https://github.com/hypjudy/Decouple-SSAD}.
Next, we evaluate the classification and proposal branches separately by incorporating each of them with the main stream.
We then evaluate another new framework which refines the localization and classification in only one branch through deconvolution, denoted as refinement branch.
Finally, we incorporate both branches with the main stream, which is our proposed Decouple-SSAD.

Table \ref{tab:two-branch} summarizes the mAP@0.5 performances of different schemes.
As expected, both classification and proposal branches could boost the performances against the baseline, where 1\% and 2.2\% improvements are obtained respectively.
Moreover, combining both branches could further enhance the performance significantly, providing 4.6\% enhancement against the baseline.
The results indicate the advantage of leveraging deeper semantic information for refining the localization and classification results. 
It can be observed that adding the refinement branch could also improve the performance by 2.8\%. However, such improvement is much lower than the result produced by Decouple-SSAD.
Though both methods involve utilization of semantics from deeper layers, they are fundamentally different in the way that the former simply mixes up the localization and classification components, while our Decouple-SSAD decouples them to learn the specialized features useful for optimizing the performance of each component.
As indicated by our results, such divide-and-conquer mechanism could achieve better performance, which verifies the design of Decouple-SSAD.

\begin{table}[t]
\centering
\caption{mAP comparisons on THUMOS14 with various IoU threshold. The last two rows denote the results of our method pre-trained on UCF101 and Kinetics dataset respectively.} 
\label{tab:results}
\begin{adjustbox}{width=0.48\textwidth}
\small
\begin{tabular}{ c|c|c c c c c } 
\Xhline{2\arrayrulewidth}
\multirow{2}{*}{Stage} & \multirow{2}{*}{Method} & \multicolumn{5}{c}{mAP@IoU (\%)} \\ 
& & 0.3 & 0.4 & 0.5 & 0.6 & 0.7 \\
\hline
\multirow{8}{*}{Two-stage} 
& SCNN~\cite{shou2016temporal} & 36.3 & 28.7 & 19.0 & 10.3 & 5.3 \\ 
& SST~\cite{buch2017sst} & 41.2 & 31.5 & 20.0 & 10.9 & 4.7 \\ 
& CDC~\cite{shou2017cdc} & 40.1 & 29.4 & 23.3 & 13.1 & 7.9 \\ 
& TURN~\cite{Gao_2017_ICCV} & 46.3 & 35.5 & 24.5 & 14.1 & 6.3 \\ 
& TCN~\cite{dai2017temporal} & - & 33.3 & 25.6 & 15.9 & 9.0 \\ 
& R-C3D~\cite{xu2017r} & 44.8 & 35.6 & 28.9 & 19.1 & 9.3 \\ 
& SSN~\cite{zhao2017temporal} & 51.9 & 41.0 & 29.8 & 19.6 & 10.7 \\ 
& CBR~\cite{gao2017cascaded} & 50.1 & 41.3 & 31.0 & 19.1 & 9.9 \\ 
& BSN~\cite{lin2018bsn} & 53.5 & 45.0 & 36.9 & 28.4 & \textbf{20.0} \\ 
\hline
\multirow{4}{*}{One-stage} 
& Yeung et al.~\cite{yeung2016end} & 36.0 & 26.4 & 17.1 & - & - \\ 
& SSAD~\cite{lin2017single} & 43.0 & 35.0 & 24.6 & 15.4 & 7.7 \\ 
& SS-TAD~\cite{sstad_buch_bmvc17} & 45.7 & - & 29.2 & - & 9.6 \\ \cline{2-7}
& Ours-UCF101 & 49.9 & 44.4 & 35.8 & 24.3 & 13.6 \\ 
& Ours-Kinetics & \textbf{60.2} & \textbf{54.1} & \textbf{44.2} & \textbf{32.3} & 19.1 \\ 
\Xhline{2\arrayrulewidth}
\end{tabular}
\end{adjustbox}
\end{table} 

\subsection{Comparisons with State-of-the-Art}
We compare with several state-of-the-art techniques of action detection on THUMOS14 in Table \ref{tab:results}.
The results indicate that Decouple-SSAD exhibits better performances than other one-stage methods and is comparable with the state-of-the-art two-stage approaches.
In particular, Decouple-SSAD achieves 35.8\% of mAP@0.5, which outperforms the best one-stage competitor SS-TAD by 6.6\% and is close to the best two-stage approach BSN (36.9\%).
In addition, when employing the Kinetics pre-trained feature extractor, we could achieve much better performance.
The results again validate our idea of decoupling the localization and classification to learn more specialized features.
Figure \ref{fig:visualization} showcases the detection results of three methods on a video from THUMOS14.

\section{Conclusions}
We have presented Decoupled Single Shot temporal Action Detection (Decouple-SSAD), which explores the separable localization and classification mechanism in one-stage action detection framework.
In particular, we study the problem of learning specialized features useful for optimizing each subtask through two separate branches.
Each branch produces a set of action anchor layers by applying deconvolution to the feature maps of the main stream.
In addition, high-level semantic information from deeper layers is incorporated to enhance the feature representations. The whole framework is trained in an end-to-end manner.  
More remarkably, we achieve state-of-the-art performance on the standard THUMOS14 benchmark.

\section{Acknowledgement}
This work is supported by the National Key R\&D Program of China under Grant NO.2018YFB0203904, the National Natural Science Foundation of China under Grant NO.U1611261 and the Program for Guangdong Introducing Innovative and Enterpreneurial Teams under Grant NO.2016ZT06D211.

\bibliographystyle{IEEEbib}
{\small \bibliography{icme2019template}}

\end{document}